# The Algebra of Utility Inference

Ali E. Abbas  (Work in Progress)

*Department of Management Science and Engineering, Stanford University, Stanford, Ca, 94305*

**Abstract.** Richard Cox [1] set the axiomatic foundations of probable inference and the algebra of propositions. He showed that consistency within these axioms requires certain rules for updating belief. In this paper we use the analogy between probability and utility introduced in [2] to propose an axiomatic foundation for utility inference and the algebra of preferences. We show that consistency within these axioms requires certain rules for updating preference. We discuss a class of utility functions that stems from the axioms of utility inference and show that this class is the basic building block for any general multiattribute utility function. We use this class of utility functions together with the algebra of preferences to construct utility functions represented by logical operations on the attributes.

## INTRODUCTION

It was for long doubted if there are principles of probable inference that are valid over the different schools of probability. Venn posed many such doubts in his book "Logic Of Choice". The following is an example taken from Venn's book.

"In every case in which we extend our inferences by Induction or Analogy or depending upon the witness of others, or trust to our own memory of the past, or come to a conclusion through conflicting arguments, or even make a long and complicated deduction by mathematics or logic, we have a result of which we can scarcely feel as certain as of the premises from which it was obtained. In all these cases then we are conscious of varying quantities of belief, but are the laws to which belief is produced and varied the same? If they cannot be reduced to one harmonious scheme, if in fact they can be reduced to nothing but a number of different schemes each with its own body of laws and rules, then it is vain to endeavour to force them into one science."

Richard Cox [1] sought to answer Venn's criticism. Building on the algebra of propositions, he proposed two axioms of probable inference, which led to the conclusion that the rules by which belief is updated can in fact be reduced to one harmonious scheme. Cox thought of probability as a degree of belief, and showed that Bayes' rule provides a unique non-trivial method of inference for his axioms. Cox thereby answered Venn's criticism.

In this paper we start with a related question about the algebra of preferences and the rules of utility inference. In every case in which we extend our preferences over the attributes of a given situation, we are conscious of our preference over each of the individual attributes and assign a utility function over each attribute. Given varying degrees of some attributes, are the laws by which preference is varied over the remaining attributes the same?" This question asks if there is a notion of utility inference analogous to probability inference. Is there a logical way in which preference is assigned and updated over some attributes when we are guaranteed certain amounts of the others? Do we have a unifying method to express our preferences in terms of logical operators of attributes and, furthermore, to update preferences? This paper attempts to answer these question and presents the algebra of preferences, and an axiomatic derivation for the notion of utility inference that parallel's Cox's axiomatic derivation for probable inference.

Let us first provide some notation and algebra for describing prospects of a decision situation, in terms of their attributes (which may be single or multiple).

## ALGEBRA AND NOTATION

In this section we present the notation and algebra needed to describe prospects of a decision situation in terms of the attributes involved. Let us start with the case of a prospect with a single attribute, $X$. For example, the single attribute may be money, ambient temperature, humidity, or market share of a company. We assume that the prospects are ordered such that $X_{min}$ is a least preferred prospect and $X_{max}$ is most preferred at a given state of preference, $\Psi$.

The state of preference, $\Psi$, summarizes our background state, knowledge, and preference for the attributes at a given epoch in time. $\Psi$ can change by receiving information about some of the attributes, receiving varying degrees of some attributes, or even as time passes by. For example our risk aversion for money may change with age or with wealth, our preferences for a skydiving experience may change with age, and our preferences for flying small planes may change if we receive information about their accident rates.

Utility assignment produces a higher utility value for prospects that are more preferred at a given state of preference, $\Psi$. Thus if the prospects are ordered in a ranked list, the utility assignment is a transformation from a preference order to a cardinal utility value which is a real number.

We note that the assignment of a single real number for the utility value of each prospect does not go without descriptive controversy. By representing preferences with a single real number, we are implicitly assuming that the preferences of any two prospects are comparable. For example, given two prospects, $X_A$ and $X_B$, and a state of preference, $\Psi$, either $X_A$ and $X_B$, are equally preferred, or one is preferred to the other. This also leaves no grounds for uncertainty about utility values: if a range of utility values is assigned to a single prospect, it may result in overlap with a range of utility values for a less preferred prospect and hence we become "money pumps" (we would choose prospect $X_A$ over prospect $X_B$, but also prospect $X_B$ over prospect $X_A$, and be willing to pay money to move from one prospect to the other). In an analogy with probability, Jaynes [3] argued for universal comparability of the plausibility of propositions. In this case we also argue for universal comparability of the preference of prospects.

For any two prospects, $X_A$ and $X_B$, in a ranked list, we thus have

$$X_A \mid \Psi \succ X_B \mid \Psi \Rightarrow U(X_A \mid \Psi) > U(X_B \mid \Psi) \qquad (1)$$

where the operator , $\succ$ , means "is strictly preferred to".

The preference ordering relation and its correspondence with the assigned utility values is shown in figure 1.

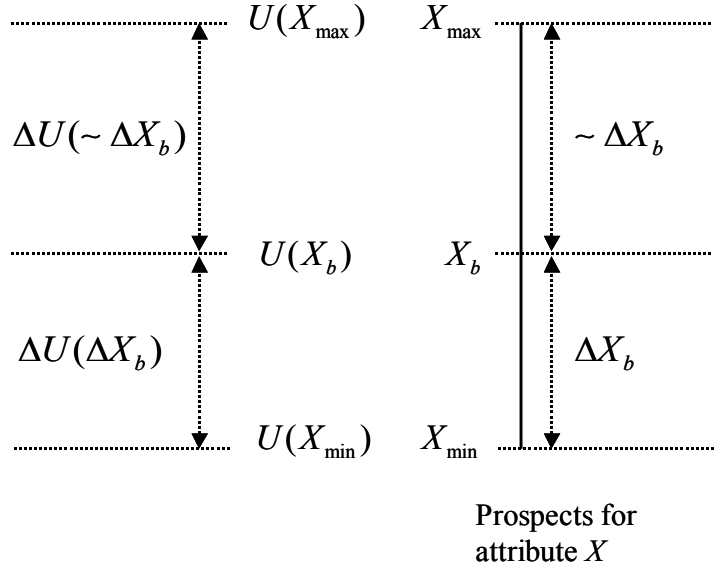

**FIGURE 1.** Utility assignment for ordered prospects.

In figure 1, the symbol $X_b$ refers to a certain value of a prospect on the ranked list. We will refer to the prospects included between $X_b$ and $X_{min}$ as the domain of $X_b$, $\Delta X_b$. This domain defines a set of prospects bounded by $X_b$ and $X_{min}$. For shorthand, we will now refer to the domain $\Delta X_b$ as simply $X_b$. We also refer to the prospects that do not lie in the domain $X_b$ as the complement of $X_b$, and will give it the notation $\sim X_b$. The complement of $X_b$ is thus a domain of prospects included in the domain of $X_b$ and $X_{max}$. The complement of a domain of prospects defines the remaining prospects that are not included in its domain.

Note that if we are indifferent between two prospects, then they have the same order on the ranked list and, as a consequence, have the same domain. They also have the same utility value assigned to them. Conversely, any two prospects that have the same domain will have the same preference order and the same utility values.

Note also that the complement of the complement of a domain of prospects is thus equal to the domain itself. I.e.

$$\sim\sim X_b = X_b \qquad (2)$$

Since $X_b$ dominates all other prospects in its domain we will refer to the difference $U(X_b) - U(X_{min})$ as the maximum utility increment of the prospects in the domain of $X_b$, and describe it using the shorthand notation $U(X_b)$. The higher the value of $U(X_b)$, the more preferred is the prospect $X_b$ and the further it is on the list from the worst prospect, $X_{\min}$. Similarly, we will refer to the difference $U(X_{\max}) - U(X_b)$ as the maximum utility increment of prospects in the complement of $X_b$, and describe it using the notation $U(\sim X_b)$. The higher the value of $U(\sim X_b)$, the less preferred is the prospect $X_b$ and the further it is on the list from the best prospect, $X_{\max}$.

## THE FIRST AXIOM: COMPLEMENTARITY

Now we are ready for the first axiom of utility inference, which we state as follows. The utility increment of a domain of a prospect, $U(X_b)$, at a given state of preference, should tell us about the utility increment of its complementary domain, $U(\sim X_b)$, at the same state of preference. In other words, as the utility value of a prospect gets higher, we would like it to tell us the degree by which the prospect gets further from the worst prospect in the domain and, at the same time, the degree by which it gets closer to the best prospect.

This can be written in mathematical form for a one-attribute prospect as

$$U(X_b \mid \Psi) - U(X_{\min} \mid \Psi) = S(U(X_{\max} \mid \Psi) - U(X_b \mid \Psi)) \qquad (3)$$

where $S$ is a continuous monotonic function.

We can write equation (3) in more concise form as:

$$U(X_b \mid \Psi) = S[U(\sim X_b \mid \Psi)] \qquad (4)$$

The utility increment of a prospect at a given state of preference determines the utility increment of its complement given the same state of preference. This axiom also applies to prospects, which have more than one attribute (as we shall discuss in the next section).

## PROSPECTS WITH MULTIPLE ATTRIBUTES

So far we have focused on prospects with one attribute. We can now extend our analysis further and consider prospects of a decision situation that have multiple attributes. Let us start with the case of two attributes. We will use the term $(X, Y)$ to represent a prospect that has two attributes, $X$ and $Y$, where $X \in [X_{\min}, X_{\max}]$ and $Y \in [Y_{\min}, Y_{\max}]$. Our analysis will refer to situations where we can order the attributes such that $(X_{\min}, Y_{\min})$ is the least preferred prospect and $(X_{\max}, Y_{\max})$ is most preferred.

Furthermore we will assume that the values of $(X,Y)$ are arranged such that for any $X_0, X_1, Y_0, Y_1$ we have

$$X_1 > X_0 \Rightarrow (X_1, Y) \mid \Psi \succ (X_0, Y) \mid \Psi, \forall Y \text{ and } Y_1 > Y_0 \Rightarrow (X, Y_1) \mid \Psi \succ (X, Y_0) \mid \Psi, \ \forall X \ (5)$$

We consider a prospect in the two-attribute case as the conjunction of two values, one for each of the attributes, using the notation, $X_b.Y_c$; to represent level $b$ of attribute $X$ and level $c$ of attribute $Y$.

When only one level of an attribute is specified in the multiattribute case, we assume all other attributes are set at their maximum values. I.e. in the two-attribute case, $X_b$ will refer to the prospect $X_b Y_{\max}$ represented by the shaded region below.

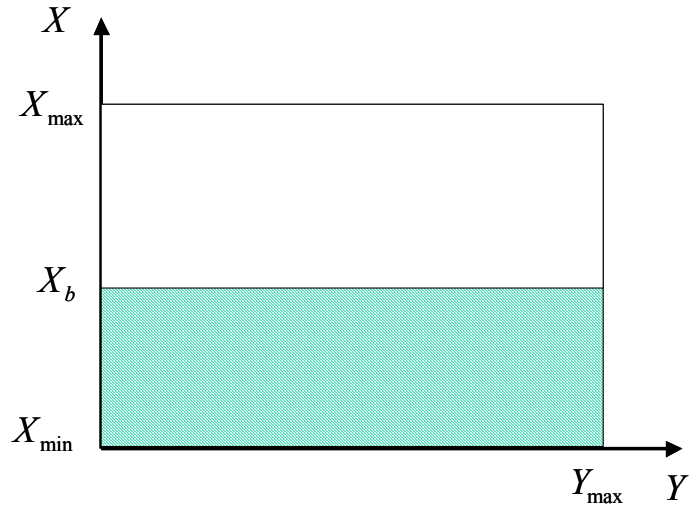

**Figure 2.** One-attribute specification in the two-attribute case.

By the construction of equation (5), the top right corner of any rectangular region dominates the remaining prospects in that region at a given state of preference. The maximum utility increment over this region thus refers to the difference in utility values between the top-right prospect and the bottom-left prospect. If the utility value of the bottom-left prospect is set equal to zero, then the utility increment is in fact equal to the utility value of the top-right prospect.

## The Conjunction Operation

The conjunction operation is the intersection of the domain of two prospects. For example, the conjunction of two prospects $X_b.Y_{\max}$ and $X_{\max} Y_c$ is the shaded region shown in figure 3.

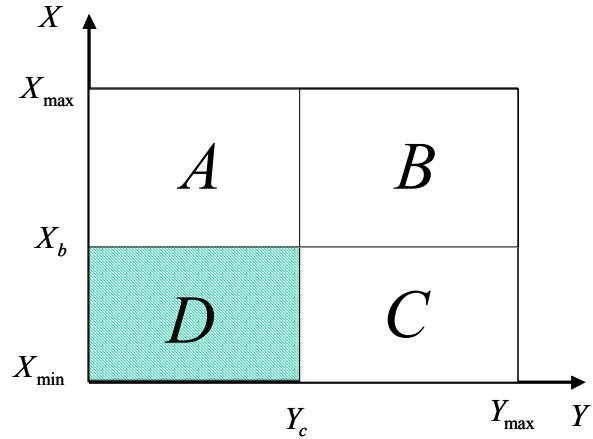

**Figure 3.** Domain of $X_b.Y_c$ obtained from the conjunction of $X_b.Y_{max}$ and $X_{max}Y_c$

The conjunction of two attributes $X_b$ and $Y_c$ forms a prospect that guarantees level $X_b$ of attribute $X$ and level $Y_c$ of attribute $Y$. We can express our preferences as the conjunction of two attributes using the framework: "we are interested in $X$ AND $Y$". This means we are interested in both of them jointly. . For example, one might say, "my values are family AND career". One of the attributes alone will not suffice no matter what level it is set to.

We note that the conjunction of two attributes forms a symmetric relationship, since it is merely a description of the level of each of the attributes of the prospect or of the values of the decision maker. For example, consider the case of two attributes, revenue AND market share that may describe values of a company. We can describe the prospect by stating the revenue first and then the market share, or we can describe the same prospect by stating its market share first, then the revenue. The order of the description is thus irrelevant and the conjunction of two attributes forms a symmetric relationship, which we write as

$$X_b.Y_c = Y_c.X_b \qquad (6)$$

Since the order of the conjunction relations is immaterial, then parentheses are also unnecessary in describing the prospects. From this result we can write for three attributes

$$(X_b.Y_c).Z_d = X_b.(Y_c.Z_d) = X_b.Y_c.Z_d \qquad (7)$$

Equation (7) demonstrates the associativity of the conjunction of three attributes. As we shall see later, this property has important consequences for the utility function of the conjunction of attributes. It is also clear that the expression $X_b.X_b$ describes the prospect by a statement about the attribute $X_b$ and that the statement is repeated twice. For example, a prospect defined by revenue equal to \$ 3 million, and revenue equal to $ 3 million (repeated twice) is the prospect of revenue equal to $ 3 million. Based on this, the conjunction of an attribute with itself produces the same attribute

$$X_b.X_b = X_b \tag{8}$$

## The Disjunction Operation

The disjunction operation is the union of the domain of two prospects. We will give this new domain the notation

$$X_b \vee Y_c \tag{9}$$

We can express our preferences as the disjunction of two attributes using the framework: "we are interested in $X$ OR $Y$". For example, let us consider that a company is interested in the net present value of profit. The attributes involved are profit in the first year of operation and the discounted profit in the second year. The utility of the company for the net present value is derived from either attribute individually or from both of them together. In other words, our preference for the net present value is the same for equal values of net present value whether it comes all from the first year's profit, all from the second year's profit or a mixture of profit in both years that yields the same net present value. The disjunction operator implies the inclusive OR meaning of the attributes and allows us to choose the "rectangular" domain we prefer. Figure 4 shows the domain described by the disjunction of the two attributes $X_b.Y_{max}$ and $X_{max}Y_c$.

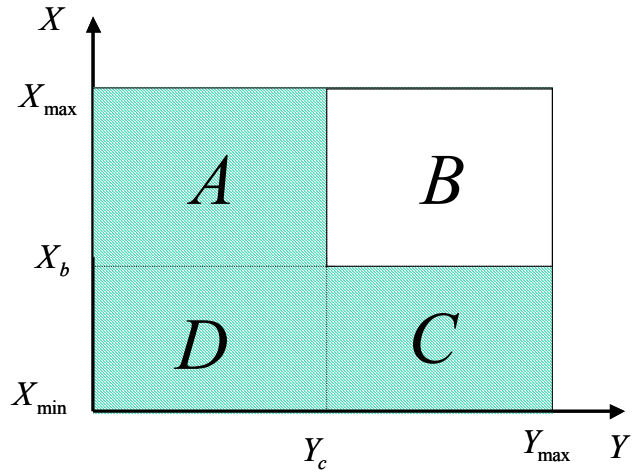

**Figure 4.** Domain of $X_b \vee Y_c$ obtained from the disjunction of $X_b.Y_{max}$ and $X_{max}Y_c$

The disjunction of two attributes is a symmetric relation since it is the same prospect but described in a different order. We can thus write

$$X_b \vee Y_c = Y_c \vee X_b \tag{10}$$

Similarly, the disjunction of three attributes is associative,

$$(X_b \vee Y_c) \vee Z_d = X_b \vee (Y_c \vee Z_d) = X_b \vee Y_c \vee Z_d \qquad (11)$$

Finally, the disjunction of an attribute with itself yields the same attribute,

$$X_b \vee X_b = X_b \qquad (12)$$

## THE ALGEBRA OF PREFERENCES

Let us now consider prospects that are formed by more than just the conjunction or disjunction of two attributes. This enables us to express more complex values in terms of logical operations of the attributes. For example, let us discuss the complement of a domain of two conjoined attributes. I.e. we want the domain of $\sim (X_b.Y_c)$, which is the non-shaded regions in figure 3 (the areas A, B, and C). Note that this domain can be obtained by the disjunction of $(\sim X_b \vee \sim Y_c)$. The region defined by the two areas A and B represents the attribute $\sim X_b$, and the region defined by the areas B and C represents the attribute $\sim Y_c$. From the equivalence of domains, we have the identity

$$\sim (X_b.Y_c) = (\sim X_b \vee \sim Y_c) \qquad (13)$$

As we have discussed earlier, equivalence in domains of prospects implies equivalence in the maximum utility increments across their domains. For example, we can now write our first axiom of utility inference for the conjunction of two attributes as:

$$U(X_b.Y_c \mid \Psi) = S[U(\sim (X_b.Y_c) \mid \Psi)] = S[U((\sim X_b \vee \sim Y_c) \mid \Psi)] \qquad (14)$$

We can also verify De Morgan's inequality for attributes using the shaded areas to verify that the domain of the prospects of each side are equivalent

$$\sim (X_b \vee Y_c) = \sim X_b. \sim Y_c \qquad (15)$$

The algebra of preferences presents us with a new framework for expressing our values. For example, a company may be interested in the motivation of its employees AND in maximizing its profit AND in not having lay-offs; alternatively, it may be interested in the motivation of its employees AND in not having lay-offs OR in maximizing its profit. As we shall see, each of these statements will produce a different multiattribute utility function.

We summarize our expressions for the algebra of preferences and provide more identities that can be verified using their equivalent domains in table 1. Note that the two equations in each row are the duals to each other and can be obtained by replacing the signs $.$ and $\vee$ (except for the first equation which is repeated twice for completeness).

| | |
|---|---|
| $\sim\sim X_b = X_b$ | $\sim\sim X_b = X_b$ |
| $X_b.X_b = X_b$ | $X_b \vee X_b = X_b$ |
| $X_b.Y_c = Y_c.X_b$ | $X_b \vee Y_c = Y_c \vee X_b$ |
| $\sim(X_b.Y_c) = \sim X_b \vee \sim Y_c$ | $\sim(X_b \vee Y_c) = \sim X_b . \sim Y_c$ |
| $(X_b.Y_c).Z_a = X_b.(Y_c.Z_a) = X_b.Y_c.Z_a$ | $(X_b \vee Y_c) \vee Z_a = X_b \vee (Y_c \vee Z_a) = X_b \vee Y_c \vee Z_a$ |
| $(X_b \vee Y_c).Z_a = (X_b.Z_a) \vee (Y_c.Z_a)$ | $(X_b.Y_c) \vee Z_a = (X_b \vee Z_a).(Y_c \vee Z_a)$ |
| $(X_b \vee Y_c).X_b = X_b$ | $(X_b.Y_c) \vee X_b = X_b$ |
| $(X_b \vee \sim X_b).Y_c = Y_c$ | $(X_b. \sim X_b) \vee Y_c = Y_c$ |
| $X_b \vee \sim X_b \vee Y_c = X_b \vee \sim X_b$ | $X_b. \sim X_b.Y_c = X_b. \sim X_b$ |

**Table 1.** The algebra of preferences.

Now we are ready for our second axiom of utility inference.

## THE SECOND AXIOM: CONJUNCTION

The second axiom relates the utility of the conjunction of two attributes to the utility and conditional utility functions of the individual attributes. It can be stated as follows: the maximum utility increment of the domain of two conjoined attributes is determined by their separate utility increments; one on the given state of preferences, and the other on this state of preferences with the additional assumption that the increment in the first is guaranteed. The intuitive rationale for this axiom is as follows. If we have a prospect $X_b.Y_c$, then it is necessary that we have level $X_b$ of attribute $X$. Thus the utility $U(X_b)$ should be involved. If in addition, we also have level $Y_c$ of attribute $Y$, then the utility $U(Y_c \mid X_b)$ is also needed. Alternatively, we can think that it is necessary that we have level $Y_c$ of attribute $Y$. Thus the utility $U(Y_c)$ should be involved. If in addition, we also have level $X_b$ of attribute $X$, then the utility $U(X_b \mid Y_c)$ is also needed.

The second axiom can be expressed in mathematical form as follows:

$$U(X_b.Y_c \mid \Psi) = F[U(X_b \mid \Psi), U(Y_c \mid X_b, \Psi)]$$
$$= F[U(Y_c \mid \Psi), U(X_b \mid Y_c, \Psi)] \tag{16}$$

where $F$ is a continuous monotonic function.

We note that some forms of multiattribute utility functions may generalize this second axiom, where the utility functions $U(X_b)$ and $U(Y_c \mid X_b)$ may be scaled by some functions of x and y. In this case, the second axiom can be generalized into

$$U(X_b.Y_c \mid \Psi) = F[a(y)U(X_b \mid \Psi) + b(y), c(x)U(Y_c \mid X_b, \Psi) + d(x)]$$
$$= F[c(x)U(Y_c \mid \Psi) + d(x), a(y)U(X_b \mid Y_c, \Psi) + b(y)] \tag{17}$$

However, we will focus on the form provided by equation (16) as it simplifies the analysis and furthermore, as, we shall see, the utility functions obtained by this axiom are the basic building blocks for any general multiattribute utility function.

The use of the pair of utility functions $U(X_b)$ and $U(Y_c \mid X_b)$ in the first line or the pair $U(Y_c)$ and $U(Y_c \mid X_b)$ in the second line of equation (16) poses the question of whether there are other combinations of for the functional form of (16). Tribus [4] posed a similar question for probability and sought the full range of exhaustive arguments of the functional equation (16). He showed that all but the two shown above lead to inconsistencies in logical thought. In an analogous form, Tribus' work can apply to utility functions so we will not elaborate on the full details of his proof in this paper. However, as an example of the analogy with his conclusions, let us examine some alternative forms of equation (16).

We might suppose that

$$U(X_b.Y_c \mid \Psi) = F[U(X_b \mid \Psi), U(Y_c \mid \Psi)] \tag{18}$$

be a permissible form. But we can show that no relation of this form could satisfy our qualitative conditions of the second axiom. The reason is that the level of attribute $X_b$ can have a very high preference given our state of preferences, $Y_c$ can also have a high preference given our state of preferences, however, the conjunction of both $X_b.Y_c$ can still be highly preferred or not preferred at all. For example, we may have a high preference for listening to classical music, and we may also have a high preference for watching action movies, however, we may have a very little preference for doing both simultaneously. We would have no way of taking such conditional influences into account with a formula like (18).

Associativity of the conjunction operation discussed earlier provides great consequences for the associativity of the function, $F$, where it limits the range of functions it can take on.

If we write equation (16) for the case of three attributes, we have

$$U(X_b.Y_c.Z_a \mid \Psi) = U(X_b.(Y_c.Z_a) \mid \Psi)$$
$$= F[U(X_b \mid (Y_c.Z_a), \Psi), U(Y_c.Z_a \mid \Psi)] \qquad (19)$$
$$= F[U(X_b \mid (Y_c.Z_a), \Psi), F[U(Y_c \mid Z_a, \Psi), U(Z_a \mid \Psi)]]$$

Alternatively, we can group the attributes in a different way to get

$$U(X_b.Y_c.Z_a \mid \Psi) = U((X_b.Y_c).Z_a \mid \Psi)$$
$$= F[U((X_b.Y_c) \mid Z_a, \Psi), U(Z_a \mid \Psi)] \qquad (20)$$
$$= F[F[U(X_b \mid Y_c, Z_a, \Psi), U(Y_c \mid Z_a, \Psi)], U(Z_a \mid \Psi)]$$

From the associativity of the conjunction, equations (19) and (20) must be equal so we have

$$F[x, F[y, z]] = F[F[x, y], z] \qquad (21)$$

The functional equations (4) and (21) are analogous to Cox's equations for probability, and have a unique non-trivial solution described by the following equations

$$U(X_b \mid \Psi) + U(\sim X_b \mid \Psi) = 1 \qquad (22)$$

$$U(X_b.Y_c \mid \Psi) = U(X_b \mid \Psi).U(Y_c \mid X_b, \Psi)$$
$$= U(Y_c \mid \Psi).U(X_b \mid Y_c, \Psi) \qquad (23)$$

Equations (22) and (23) form the basis for the assignment of a utility function for the conjunction of two attributes. They are also sufficient to determine the utility function for any logical operation of attributes.

The commutativity of the conjunction operator derived earlier, results in the two forms of equation (23) and allows for the expression

$$U(Y_c \mid X_b, \Psi) = \frac{U(Y_c \mid \Psi)U(X_b \mid Y_c, \Psi)}{U(X_b \mid \Psi)} \qquad (24)$$

Equation (24) forms the basis of utility inference and explains how the utility function over one attribute changes for different values of the others.

We now define the condition of utility independence between attributes $X$ and $Y$ when the conditional utility function of $X$ is independent of $Y$, or alternatively, the conditional utility function of $Y$ is independent of $X$. I.e.

$$U(X \mid Y, \Psi) = U(X \mid \Psi) \qquad (25)$$

Substituting equation (25) into equation (24), we find that this condition also implies

$$U(Y \mid X, \Psi) = U(Y \mid \Psi) \tag{26}$$

Equations (25) and (26) show that utility independence is a symmetric relationship. Mathematical manipulations of equations (22) and (23), as well as the algebra of preferences, allows for the deduction of the utility of the disjunction of two attributes as follows. From equation (22) we have

$$U(X_b \vee Y_c \mid \Psi) = 1 - U(\sim (X_b \vee Y_c) \mid \Psi) \tag{27}$$

From equation (15) we have

$$U(\sim (X_b \vee Y_c) \mid \Psi) = 1 - U((\sim X_b).(\sim Y_c) \mid \Psi) \tag{28}$$

From equation (23), we have

$$
\begin{aligned}
U(X_b \vee Y_c \mid \Psi) &= 1 - U(\sim X_b \mid \Psi)U(\sim Y_c \mid \sim X_b, \Psi) \\
&= 1 - U(\sim X_b \mid \Psi)(1 - U(Y_c \mid \sim X_b, \Psi)) \\
&= 1 - U(\sim X_b \mid \Psi) + U(\sim X_b.Y_c \mid \Psi) \\
&= U(X_b \mid \Psi) + U(Y_c \mid \Psi)U(\sim X_b \mid Y_c, \Psi) \\
&= U(X_b \mid \Psi) + U(Y_c \mid \Psi)(1 - U(X_b \mid Y_c, \Psi)) \\
&= U(X_b \mid \Psi) + U(Y_c \mid \Psi) - U(X_b.Y_c \mid \Psi)
\end{aligned} \tag{29}
$$

Equation (29) provides a normative basis for assigning utility values for the disjunction of two attributes. An interesting observation we can deduce from (29) is that the disjunction of utility independent attributes has a multilinear form, which agrees with the multilinear utility function proposed in [5].

In the next section, we show how we can use the algebra of preferences and the rules of utility inference to model more general classes of multiattribute utility functions.

## LOGICAL OPERATIONS ON ATRRIBUTES

The construction of the utility value for the conjunction of two attributes provided by equation (16) assumes a value of zero when any one of the attributes is set at its minimum value. I.e.

$$U(x_{\min}, y) = U(x, y_{\min}) = 0 \qquad \forall x \in [x_{\min}, x_{\max}], \forall y \in [y_{\min}, y_{\max}] \tag{30}$$

We previously defined this type of multiattribute utility functions, that is the conjunction of attributes, as Attribute Dominance Utility [6], [7], since any attribute set at a minimum value dominates the utility function and brings it to a minimum.

In many cases that arise in practice, our preferences are described by more than just the conjunction of attributes. For example, we may be interested in the disjunction of attributes. One typical example may be the utility function for the success of a project as measured by levels of market share and/or revenue. Both of these attributes individually or combined may drive our preferences for the prospect. In this case we can use the algebra of prospects discussed earlier to construct a utility function as:

$$U(Y_c \vee X_b \mid \Psi) = U(Y_c \mid \Psi) + U(X_b \mid \Psi) - U(Y_c.X_b \mid \Psi) \tag{31}$$

We can also use Bayes' rule for utility inference in this case. For example,

$$U(Y_c \vee X_b \mid Z_a, \Psi) = \frac{U(Y_c.Z_a \mid \Psi) + U(X_b.Z_a \mid \Psi) - U(Y_c.X_b.Z_a \mid \Psi)}{U(Z_a \mid \Psi)} \tag{32}$$

The algebra of preferences assigned to utility functions allows us to think of our preferences more generally in terms of logical operations on the attributes rather than just their conjunction.

In the following example, we show how the extension of attribute dominance utility, to include logical operations of attributes, explains many general formulations of classes of multiattribute utility functions.

## Example: Utility function for disjunction of attributes

Let us consider a corporation that assigns a multiattribute utility function for profit. The attributes associated with the company's utility function are profit in the first year of operation and profit in the second year. The company has an exponential utility function and is risk averse for the net present value of profit, with a risk aversion coefficient, $\gamma$.

We can think of the construction of this utility function using the algebra of preferences that we have developed. If the company is only interested in profit, then its utility comes from either attribute $x$, or a discounted attribute $y$, or both attributes conjoined together. The company's utility is based on the disjunction of the two operators. Using the rules of the algebra of preferences, we have

$$U(x \vee y \mid \Psi) = U(x \mid \Psi) + U(y \mid \Psi) - U(x.y \mid \Psi) \tag{33}$$

The conjunction of the two attributes $U(x.y \mid \Psi)$ is their attribute dominance utility function and, for the case of exponential utility functions, is the product of the utility functions for each attribute (since the risk attitude does not depend on the wealth state, then the utility function for the second year profit is independent of the first). Therefore we can write equation (33) as

$$U(x \vee y \mid \Psi) = U(x \mid \Psi) + U(y \mid \Psi) - U(x \mid \Psi)U(x \mid \Psi) \tag{34}$$

If the company is risk averse for profit and has an exponential utility function, then consistency of assigning utility functions requires the following utility functions for each of the attributes:

$$U(x \mid \Psi) = 1 - e^{-\gamma x}, x \geq 0 \tag{35}$$

$$U(y \mid \Psi) = 1 - e^{-\gamma \beta y}, x \geq 0 \tag{36}$$

If we substitute the utility function for attributes $x$ and $y$ from (35) and (36) into equation (34), we get

$$U(x \vee y \mid \Psi) = (1 - e^{-\gamma(x)}) + (1 - e^{-\gamma(\beta y)}) - (1 - e^{-\gamma(x)})(1 - e^{-\gamma(\beta y)}), \quad x, y \geq 0.$$
$$= 1 - e^{-\gamma(x + \beta y)}, \quad x, y \geq 0. \tag{37}$$

The results of the algebra of preferences provide a new framework for thinking about the utility values of different attributes. In fact, the prospects and values can be expressed in terms of more complicated operations but we note that the terms for all the logical expressions in the final multiattribute utility function can be determined from the attribute dominance utility function (the utility function for the conjunction).

## CONCLUSIONS

In this paper we have introduced axioms of utility assignment and inference using an analogy with Cox's axiomatic approach for probability. We also developed the algebra of preferences, which describes the values in terms of logical operations on their attributes. The algebra of preferences, combined with the rules of utility inference, allows the construction of more general classes of multiattribute utility functions using logical identities as was demonstrated with the disjunction operation.

## ACKNOWLEDGEMENTS

I would like to thank Ronald Howard and Myron Tribus for introducing me to the work of Richard Cox and Edwin Jaynes.